\documentclass[sigconf,screen]{acmart}
\usepackage{listings}
\usepackage{multirow}
\usepackage[linesnumbered,ruled, boxed]{algorithm2e}
\AtBeginDocument{%
  }

\setcopyright{acmlicensed}
\copyrightyear{2018}
\acmYear{2018}
\acmDOI{XXXXXXX.XXXXXXX}

\acmConference[Conference acronym 'XX]{Make sure to enter the correct
  conference title from your rights confirmation emai}{June 03--05,
  2018}{Woodstock, NY}
\acmISBN{978-1-4503-XXXX-X/18/06}

\begin{document}

\title{Node Level Graph Autoencoder: Unified Pretraining for Textual Graph Learning}

\author{Wenbin Hu}
\authornote{Authors contributed equally to this research.}
\affiliation{%
  \institution{Department of CSE, Hong Kong University of Science and Technology}
  \city{Hong Kong}
  \country{China}}
\email{whuak@connect.ust.hk}

\author{Huihao Jing}
\authornotemark[1]
\affiliation{%
  \institution{Department of CSE, Hong Kong University of Science and Technology}
  \city{Hong Kong}
  \country{China}}
\email{hjingaa@connect.ust.hk}

\author{Qi Hu}
\authornotemark[1]
\affiliation{%
  \institution{Department of CSE, Hong Kong University of Science and Technology}
  \city{Hong Kong}
  \country{China}}
\email{qhuaf@connect.ust.hk}

\author{Haoran Li}
\affiliation{%
  \institution{Department of CSE, Hong Kong University of Science and Technology}
  \city{Hong Kong}
  \country{China}}
\email{hlibt@connect.ust.hk}

\author{Yangqiu Song}
\affiliation{%
  \institution{Department of CSE, Hong Kong University of Science and Technology}
  \city{Hong Kong}
  \country{China}}
\email{yqsong@cse.ust.hk}

\renewcommand{\shortauthors}{Trovato et al.}

\begin{abstract}
 Textual graphs are ubiquitous in real-world applications, featuring rich text information with complex relationships, which enables advanced research across various fields. Textual graph representation learning aims to generate low-dimensional feature embeddings from textual graphs that can improve the performance of downstream tasks. A high-quality feature embedding should effectively capture both the structural and the textual information in a textual graph. However, most textual graph dataset benchmarks rely on word2vec techniques to generate feature embeddings, which inherently limits their capabilities. Recent works on textual graph representation learning can be categorized into two folds: supervised and unsupervised methods. Supervised methods finetune a language model on labeled nodes, which have limited capabilities when labeled data is scarce. Unsupervised methods, on the other hand, extract feature embeddings by developing complex training pipelines. 
 To address these limitations, we propose a novel unified unsupervised learning autoencoder framework, named \textbf{Node} Level \textbf{G}raph \textbf{A}uto\textbf{E}ncoder (\textbf{NodeGAE}). We employ language models as the backbone of the autoencoder, with pretraining on text reconstruction. Additionally, we add an auxiliary loss term to make the feature embeddings aware of the local graph structure. Our method maintains simplicity in the training process and demonstrates generalizability across diverse textual graphs and downstream tasks. We evaluate our method on two core graph representation learning downstream tasks: node classification and link prediction. Comprehensive experiments demonstrate that our approach substantially enhances the performance of diverse graph neural networks (GNNs) across multiple textual graph datasets. Remarkably, a two-layer GNN can achieve a testing accuracy of 77.10\% on the ogbn-arxiv dataset. Furthermore, by ensembling with GNNs from existing SOTA methods, our method achieves a new SOTA of 78.34\%. 
\end{abstract}




\maketitle

\section{Introduction}
Textual graphs are graph-based data that incorporate textual attributes such as phrases, sentences, or documents, where each entity contains a segment of text. The rich information from textual attributes in textual graphs significantly enhances the performance across a diverse range of real-world applications, such as citation graphs~\cite{hu2021opengraphbenchmarkdatasets, wang2020MAG}, social networks~\cite{mcauley2013social, cai2017social_2}, knowledge graphs~\cite{yang2016revisitingsemisupervisedlearninggraph,speer2018KG,sap2019KG_2}, and recommendation systems~\cite{mcauley2015recsys, He_2016recsys_2}. 

\begin{figure}[t]
    \vspace{0.15in}
    \centering
    \includegraphics[width=0.47\textwidth]{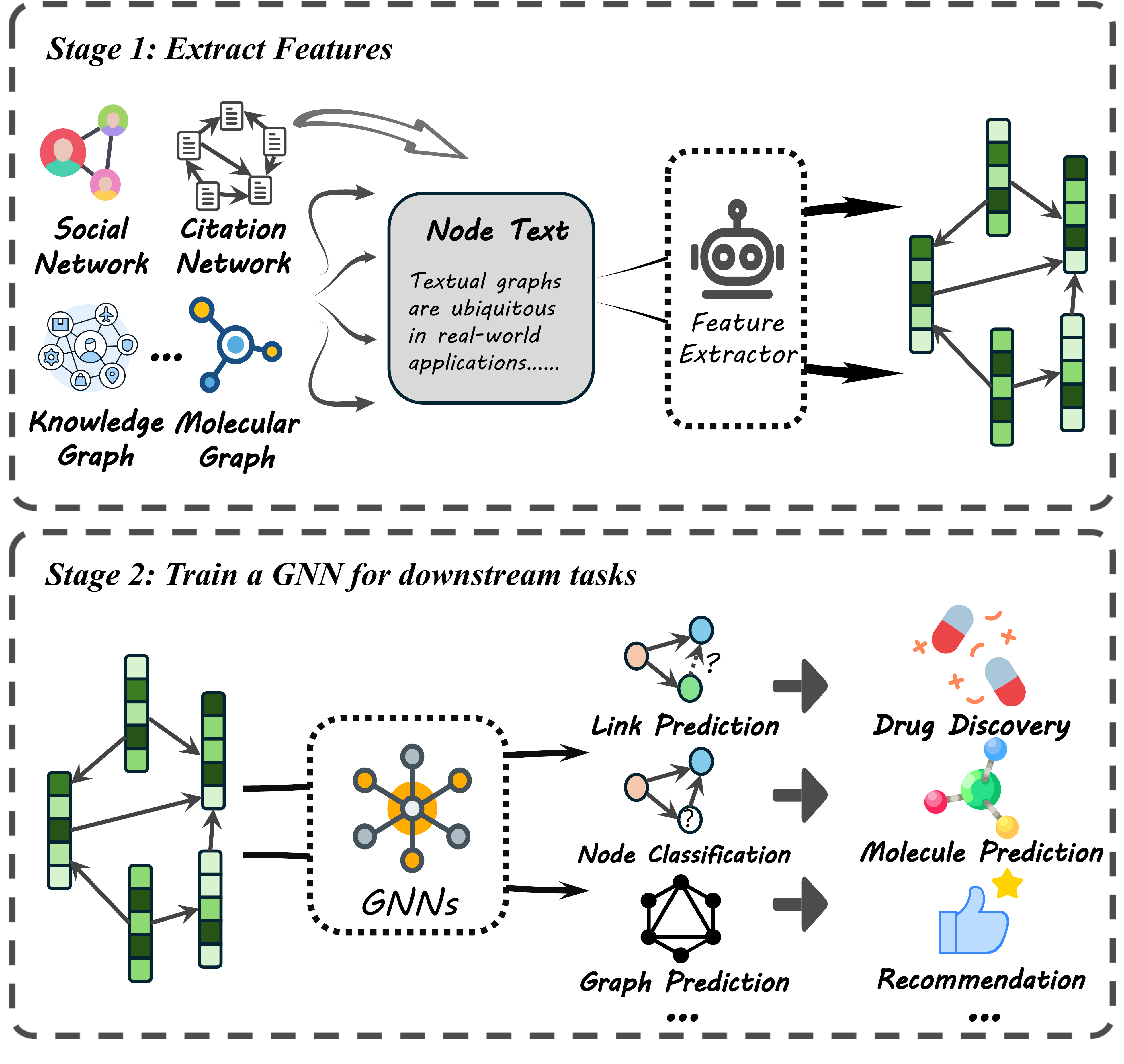}
    \vspace{-0.05in}
   \caption{The paradigm for textual graph learning.}
    \label{fig:textual_graph}
    \vspace{-0.15in}
\end{figure}

Unlike traditional Natural Language Processing (NLP) tasks, the text on the nodes in a textual graph is correlated with each other, which is important for downstream training and inference. For example, the ogbn-arxiv dataset~\cite{hu2021opengraphbenchmarkdatasets} is a citation network for academic articles, where the nodes contain the title and abstract of the corresponding article, and the edges represent citations between the articles. 
As shown in Figure \ref{fig:textual_graph}, textual graph learning typically involves two stages: 1) Extracting features from the text of the nodes, and 2) Training graph neural networks (GNNs) on the extracted node features. The second stage has been well-studied, and there are powerful GNN models available to solve it~\cite{hamilton2018graphsage, kipf2017gcn, li2022revgat}. The former stage, however, still requires more effective feature extractors, which is an active area of research in the field of textual graph representation learning.

Representation learning for textual graphs aims to create low-dimensional embeddings that can effectively represent the data points. The goal is for the feature embeddings in the low-dimensional space to capture the information from both the text and the graph structure, enabling good performance on downstream tasks. However, in most popular benchmarks, the representation embeddings are generated using techniques like skip-gram or bag-of-words~\cite{mikolov2013word2vec}, which can inherently limit the performance on downstream tasks. Recently, research has explored ways to improve the quality of feature embeddings, which can generally be categorized as supervised and unsupervised methods. For the supervised approach, SimTeG~\cite{duan2023simteg} leverages language models finetuned on the labeled nodes for feature extraction. Concurrently, TAPE~\cite{he2024tape} utilizes GPT-3.5 turbo's~\cite{ouyang2022traininglanguagemodelsfollow} text prediction and explanation capabilities to further improve performance. While the supervised method framework is simple and effective, it can be challenging to generalize to different downstream tasks. For example, in SimTeG~\cite{duan2023simteg}, the framework for node classification needs to be modified for link prediction, and the performance improvement for link prediction is not as significant. Additionally, supervised methods struggle when labeled data is scarce, as language models require a decent amount of data for training. For unsupervised methods, a representative example is GIANT~\cite{chien2022giant}, which conducts neighbor prediction for unsupervised training. However, this approach requires a complex training pipeline for unsupervised learning on textual graphs. To address the limitations of existing supervised and unsupervised methods, we propose a novel pretraining approach (NodeGAE) that leverages an autoencoder architecture. This method maintains simplicity in the training process while demonstrating generalizability across diverse textual graphs and downstream tasks.

Autoencoders have proven to be effective feature extractors~\cite{kingma2022autoencoder_cv, shen2020autoencoder_nlp}. An autoencoder is composed of two main components: an encoder and a decoder. The encoder maps the input data to a lower-dimensional latent representation, and the decoder then attempts to reconstruct the original input from this latent representation. The encoder network effectively learns a compressed representation of the input data, which can capture the most salient features or characteristics of the input, where the latent representation can serve as the extracted feature to enhance the performance of downstream tasks, such as classification or regression. Autoencoders can learn features in an unsupervised manner, without the need for the supervised signal. This can be particularly useful when labeled data is scarce or expensive to obtain, as the autoencoder can extract meaningful features from the unlabeled input data. Inspired by the autoencoder mechanism, we propose a novel node-level graph autoencoder framework (NodeGAE) for textual graph learning, with a text reconstruction objective for textual information extraction. Additionally, to capture the graph structural information, we use InfoNCE loss~\cite{oord2019infonce} to enhance the similarity for neighboring embeddings. NodeGAE can generate high-quality embeddings and achieve strong performance on downstream node classification and link prediction tasks. Our contributions are summarized as follows:
\begin{itemize}
    \item \textbf{Novel Node-Level Graph Autoencoder Architecture.} We propose a novel node-level autoencoder to enhance the performance of textual graphs. Our approach leverages text reconstruction as the unsupervised learning task, where the encoder maps the hidden node embeddings to the corresponding textual data. This allows the encoder to effectively extract and preserve valuable textual information. Furthermore, the feature embeddings extracted from the encoder are encouraged to learn structural information by utilizing the InfoNCE loss. Our method keeps the training process simple while demonstrating the ability to generalize across a diverse set of textual graphs and downstream tasks.
    \item \textbf{Comprehensive Experiment.} We conduct a comprehensive evaluation of the quality of NodeGAE's node embeddings. Our results demonstrate that NodeGAE achieves the best performance across various GNN models and datasets for both node classification and link prediction tasks. Additionally, we find that NodeGAE accelerates the convergence rate of GNNs. We further perform ablation studies to verify the effectiveness of the core components in NodeGAE. Finally, we provide insights into the text reconstruction process of NodeGAE.
    \item \textbf{SOTA Performance.} On the ogbn-arxiv dataset~\cite{hu2021opengraphbenchmarkdatasets}, NodeGAE achieves a testing accuracy of 77.10\%, which is on par with SOTA methods such as SimTeG~\cite{duan2023simteg} and TAPE~\cite{he2024tape}. Furthermore, by ensembling with GNNs from existing SOTA methods, NodeGAE is able to reach a new SOTA accuracy of 78.34\% on the ogbn-arxiv.

\end{itemize}

\begin{figure*}[!h]
    \centering
    \includegraphics[width=0.98\textwidth]{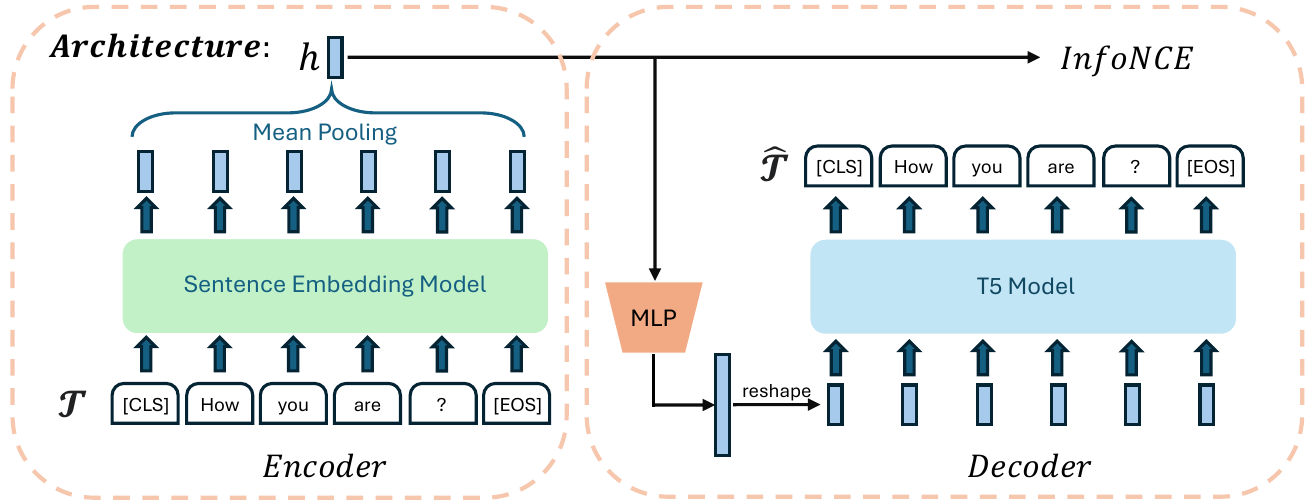}
    \vspace{-0.05in}
   \caption{Overview of the autoencoder architecture in NodeGAE. The encoder generates the latent representation $h$ as an extracted feature representation of the input text sequence $\mathcal{T}$. The decoder then reconstructs the original text sequence from the projected $h$. Moreover, the latent representation $h$ is encouraged to learn the structural information of the input text by utilizing the InfoNCE loss function.}
    \label{fig:architecture}
    \vspace{-0.15in}
\end{figure*}

\section{Related Work}
\subsection{Utilizing LMs for Textual Graph} For textual graph learning, language models (LMs) are an essential component for capturing text information. The LM+GNN paradigm has become the mainstream approach for textual graph-related tasks. Existing LM+GNN methods can be divided into two types: one-step and two-step methods. For one-step methods, the most common structure is a cascade architecture, where the LM serves as the encoder for embedding text. In the cascade structure, the LM and the GNN are trained simultaneously. TextGNN~\cite{zhu2021textgnn} first explores the cascade architecture, and AdsGNN~\cite{li2021adsgnn} further extends it by proposing edge-level information aggregation. However, such simple cascade architecture suffers from inefficiency, as it can only fit a few samples from one-hop neighbors due to memory complexity. Apart from the cascade structure, GLEM~\cite{zhao2023glem} jointly trains a LM and a GNN by passing the generated pseudo-labels to each other, formulated in an EM algorithm framework. For two-step methods, they usually train a feature extractor for feature embeddings in the first stage, and then GNNs are trained on the embeddings in the second stage. SimTeG~\cite{duan2023simteg} designs a supervised manner: finetuning a LM on labeled nodes and then using the finetuned LM as the feature extractor. It also utilizes parameter-efficient finetuning (PEFT)~\cite{hu2021lora} to mitigate overfitting. Concurrently, TAPE~\cite{he2024tape} leverages auxiliary prediction and explanation from GPT-3.5 turbo~\cite{ouyang2022traininglanguagemodelsfollow} to improve performance. For unsupervised methods, GIANT~\cite{chien2022giant} conducts neighbor prediction as the pretraining task and generates better feature embeddings compared to vanilla BERT embeddings. The two-step training strategies can effectively mitigate the problem of insufficient training of LMs, leading to higher-quality text representations. However, for existing supervised methods, it is not trivial to generalize to different downstream tasks and fails when labeled data is scarce; for unsupervised methods, the training pipeline is complex and memory-intensive. Our proposed NodeGAE provides a simple and unified pretraining framework for different downstream tasks. Since NodeGAE is on the node level, it is memory-friendly for training. \\

\vspace{-0.1in}

\subsection{Graph Pretraining Frameworks} 
Significant progress has been made in developing graph pretraining to learn expressive representations for GNNs. Several GNN pretraining frameworks have been proposed: 1) Graph Autoregressive Modeling: An autoregressive framework to perform iterative graph reconstruction. GPT-GNN~\cite{hu2020gptgnn} predicts one masked node and its edges at a time given a graph with randomly masked nodes and edges. MGSSL~\cite{zhang2021MGSSL} generates molecular graphs in an autoregressive manner. 2) Masked Components Modeling: Masking out some components in a graph and training a GNN to predict them. Hu \textit{et al.}~\cite{hu2020strategiespretraininggraphneural} propose attribute masking where some attributes on nodes or edges are masked out for prediction. GROVER~\cite{rong2020grover} masks out some subgraphs in a molecular graph to capture the contextual information. 3) Graph Contrastive Learning: Constructing a self-supervised learning objective to learn representations that capture the structural and semantic similarities between graph components. DGI~\cite{velickovic2018DGI} and InfoGraph~\cite{sun2020infograph} enhance the representation of a graph by maximizing the mutual information between the graph-level structure and subgraph-level structure. MVGRL~\cite{hassani2020mvgrl} uses node diffusion to generate augmented nodes and maximizes the mutual information between the augmented and original nodes. GRACE~\cite{zhu2020grace} and its variants~\cite{Zhu_2021, xia2022progclrethinkinghardnegative} maximize the agreement of the different two augmented views of the node representation. GraphCL~\cite{you2021graphcl} and its variants~\cite{susheel2021advance, you2021graphcontrastivelearningautomated} propose new contrastive strategies for graph pretraining. Unfortunately, these existing graph pretraining frameworks cannot be trivially adapted to textual graphs, and the modified frameworks still may not fully effectively capture both the textual and structural information. NodeGAE proposes a novel pretraining framework for textual graphs, which can capture the textual and structural information simultaneously.

\begin{figure*}[!h]
    \centering
    \includegraphics[width=0.98\textwidth]{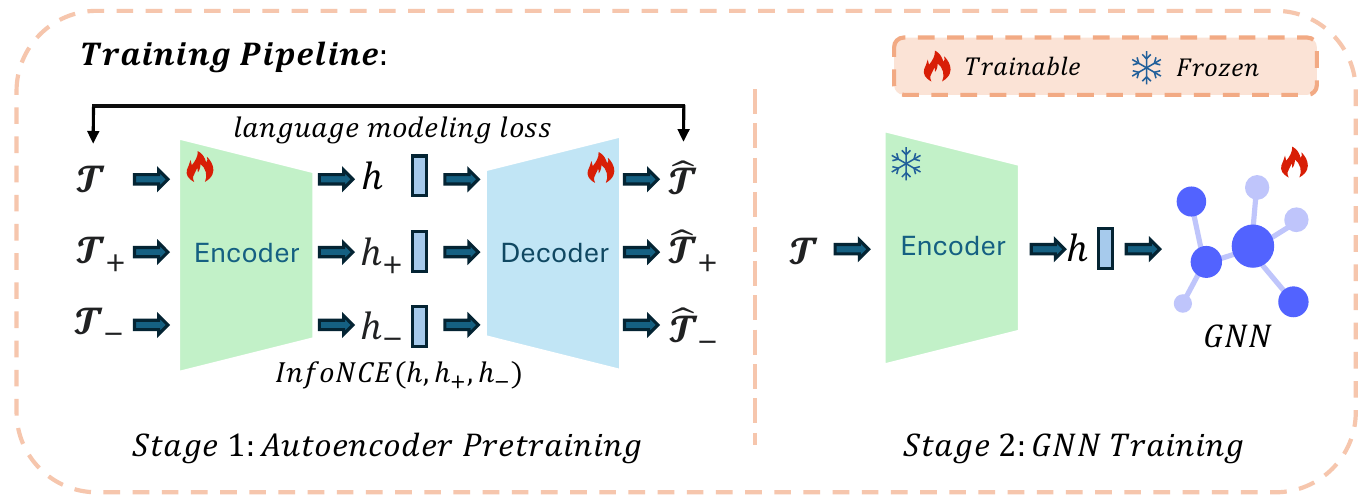}
    \vspace{-0.05in}
   \caption{Overview of NodeGAE training pipeline. The training pipeline consists of two stages. In the first stage, the autoencoder is pretrained on the text reconstruction task with the language modeling loss and the InfoNCE loss. In the second stage, a GNN is trained on the feature embeddings extracted from the frozen encoder.}
    \label{fig:training_pipeline}
    \vspace{-0.15in}
\end{figure*} 

\subsection{Autoencoders for Feature Extraction} In computer vision, the encoded features from an autoencoder can capture important visual characteristics of the input images, such as edges, textures, and shapes~\cite{kingma2022autoencoder_cv}. Similarly, in natural language processing, the encoded features from an autoencoder can capture semantic and syntactic information from the input text~\cite{shen2020autoencoder_nlp}.In graph-based learning, Graph Autoencoder (GAE) reconstructs the original graph structure, such as the adjacency matrix or the presence of edges between nodes, which serves as a graph-level autoencoder. Representatives are VGAE~\cite{kipf2016gae}, MGAE~\cite{tan2022mgae}, SIGGAE~\cite{hasanzadeh2020siggae}, and so on. Our proposed NodeGAE is a novel autoencoder architecture, where the reconstruction is performed at the node level. \\

\section{Preliminary}
In this section, we will formally define the problem we are addressing and introduce key concepts related to our proposed method. \\

\vspace{-0.1in}

\subsection{Problem Formulation} We denote the textual graph as $\mathcal{G}=(\mathcal{V}, \mathcal{E})$, where $\mathcal{V}$ and $\mathcal{E}\subseteq\mathcal{V} \times \mathcal{V} $ represent the set of nodes and the set of edges respectively. We convert the edge set $\mathcal{E}$ to an adjacency matrix $A \in \{0,1\}^{|\mathcal{V}|\times|\mathcal{V}|}$. For each node $v\in\mathcal{V}$, the node attribute is a text sequence $t\in \mathcal{T}$, where $\mathcal{T}$ is the set of textual attributes, aligning with the set of nodes $\mathcal{V}$. We focus on two fundamental problems in textual graphs: node classification and link prediction. For node classification, our goal is to build a model $\Phi: \mathcal{V} \rightarrow \mathcal{Y}$ to predict the label $y\in \mathcal{Y}$ of the node, where $\mathcal{Y} \in \mathbb{R}^{C}$ and $C$ is the number of classes. For link prediction, we aim to construct a model $\Phi: \mathcal{V}\times\mathcal{V} \rightarrow \{0,1\}$ to predict the linkage between two nodes, where $\Phi(v_i, v_j)=1$ if there exists a link between node $(v_i, v_j)$, otherwise $\Phi(v_i, v_j)=0$. \\

\vspace{-0.1in}

\subsection{GNN for Textual Graph Learning} GNN provides a unified framework to make predictions on graphs, which recently dominates the field of graph learning. Generally, GNN recursively aggregates the neighbour feature embeddings to predict the properties of nodes. For simplicity, we take the vanilla Graph Convolutional Network (GCN)~\cite{kipf2017gcn} for formulation. A GCN layer can be formulated as: $\sigma(\tilde{A}HW)$, where $\tilde{A}=D^{-\frac{1}{2}}AD^{-\frac{1}{2}}$, $D_{ii} = \sum_{j\in|\mathcal{V}|} A_{ij}, \forall i\in|\mathcal{V}|$, $H\in\mathbb{R}^{|\mathcal{V}|\times d}$ is the feature embedding matrix with the embedding dimension $d$, $W\in\mathbb{R}^{d\times d_w}$ is the model weight with the output dimension $d_w$, and $\sigma(\cdot)$ is the activation function. For node classification, a classifier can be directly appended to the final layer of the GNN to predict the class of the node; while for link prediction, a similarity function is adopted to compute the similarity score between two node embeddings. Unlike traditional graph-based tasks, the features of textual graphs do not directly construct feature embeddings for GNN training. Instead, the features are text sequences, which poses a challenge for creating feature embeddings that can capture both the text information and the graph structure information, which can be formulated as: $H = \Psi(\mathcal{T},\mathcal{V}, A).$ In most graph benchmarks~\cite{hu2021opengraphbenchmarkdatasets, yang2016revisitingsemisupervisedlearninggraph}, the feature embeddings are generated using techniques such as skip-gram or bag-of-word~\cite{mikolov2013word2vec}. Additionally, recent works have proposed new approaches for more effective feature extraction in textual graph settings~\cite{duan2023simteg, he2024tape, chien2022giant}.\\


\vspace{-0.1in}

\subsection{InfoNCE Loss}
InfoNCE loss~\cite{oord2019infonce} is a commonly used objective function for self-supervised representation learning. The core idea behind InfoNCE is to encourage the model to learn similar representations for data points that are considered 'positive' pairs, while learning distinct representations for 'negative' pairs of dissimilar data points . Formally, the InfoNCE loss can be expressed as:
\begin{equation}
L_{InfoNCE} = -\log \frac{exp(h^T h_+ / \tau)}{exp(h^T h_+ / \tau)+\sum^n_{i=1} exp(h^T h_{-i}/ \tau)},
\end{equation}
where $h\in\mathbb{R}^d$ is a $d$-dimensional latent representation of the data, $h{+}, h{-}$ are the latent representations for the positive sample $\mathcal{T_+}$ and the negative sample $\mathcal{T_-}$, $\{h_{-i}\}_{i\in\{1,..,n\}}$ is the set of negative samples, and $\tau$ is the temperature. We use InfoNCE loss to incentivize the autoencoder to learn the graph structure during the pretraining phase. \\

\vspace{-0.1in}

\begin{algorithm}[h]
\caption{NodeGAE for Node Classification}
\label{alg}
\SetKwInOut{Input}{Input}
\SetKwInOut{Output}{Output}
\SetKwInOut{Model}{Model}
\Input{$G$: textual graph with $G.X$: original feature embedding matrix, $G.V$: node list, and $G.A$: adjacency matrix; $T$: inputs\_ids for language models.}
\Output{$H$: generated feature embedding matrix from the autoencoder.}
\Model{$f\_encoder$ and $f\_decoder$: encoder and decoder, $f\_mlp$: projection MLP, $f\_gnn$: GNN model.}
\Begin{
    \For{$T,G$ in $autoencoder\_loader$}{
        $H \gets f\_encoder(T)$\;
        $\hat{T} \gets f\_decoder(f\_mlp(H))$\;
        $T\_positive \gets G.V.neighbour()$\;
        $H\_positive \gets f\_encoder(T\_positive)$\;
        $loss_1 \gets language\_modeling\_loss(\hat{T},T)$\;
        $loss_2 \gets InfoNCE\_loss(H, H\_positive)$\;
        $loss \gets loss_1 + loss_2$\;
        $loss.backward()$\;
        $autoencoder\_optimizer.step()$\;
    }
    
    $H \gets f\_encoder(T)$\;
    $G.X \gets H$\;
    
    \For{$G$ in $gnn\_loader$}{
        $\hat{Y} \gets f\_gnn(G.A,G.X)$\;
        $loss \gets CrossEntropyLoss(\hat{Y},G.Y)$\;
        $loss.backward()$\;
        $gnn\_optimizer.step()$\;
    }
}
\end{algorithm}

\vspace{-0.1in}

\section{Node Level Graph Autoencoder}
We propose a novel pretraining framework for improving textual graph learning: the Node Level Graph Autoencoder (NodeGAE). The architecture of NodeGAE is shown in Figure \ref{fig:architecture}, and the whole training pipeline is illustrated in Figure \ref{fig:training_pipeline}. Our method follows a two-stage training pipeline: 1) First, we train an autoencoder in a self-supervised manner to reconstruct the text attributes on the nodes. The encoder takes a text sequence $t_{1:M}$ with a sequence length of $M$ as the input and outputs the extracted feature embedding $h$:
\begin{equation}
h = LM_{encoder}(t_{1:M}).
\end{equation}
Then, the feature embedding is fed into the decoder to reconstruct text sequence $\hat{t}_{1:L}$ with a sequence length of $L$:
\begin{equation}
\hat{t}_{1:L} = LM_{decoder}(h).
\end{equation}
We take the language modeling loss as the text reconstruction loss:
\begin{equation}
L_{LM}(t_{1:M}) = -\sum_{i=1}^M \log p(t_i|t_{<i}).
\end{equation}
Under the self-supervised learning framework, the text reconstruction objective trains the encoder and decoder simultaneously. During this training process, the autoencoder effectively learns a mapping between the latent embeddings and the corresponding textual data. As a result, the latent embeddings can serve as a powerful representation of the data within the textual graph. 2) Then, we take the frozen encoder of the trained autoencoder as a feature extractor to obtain feature embeddings for the downstream GNN training. For the downstream tasks, we focus on node classification and link prediction.

The pseudo-code in PyTorch-style~\cite{paszke2019pytorch} for the whole training process is demonstrated in Algorithm \ref{alg}. For the limited space of the paper, the algorithm is only written for the task of node classification. \\

\vspace{-0.2in}

\subsection{Graph Structure Learning} Textual graph learning combines text learning with graph structure learning. Through text reconstruction, the autoencoder can effectively capture the semantic information of the text attributes. We also want the autoencoder to learn the local structure of each node in order to generate improved node embeddings. For graph structure learning, we leverage the InfoNCE loss~\cite{oord2019infonce} to learn the structural information. Positive samples are drawn from the node's neighbors, while negative samples come from other data points in the same batch. Furthermore, we can calculate a separate InfoNCE loss for neighbors at different hops from the node:
\begin{equation}
L_{InfoNCE} = -\sum_k \alpha_k \log \frac{exp(h^T h_{+}^{(k)} / \tau)}{exp(h^T h_{+}^{(k)} / \tau) + \sum_i exp(h^T h_{-i} / \tau)},
\label{eq:loss_neighbour}
\end{equation}
where $h_{+}^{(k)}$ is the embedding of a node from k-hop neighbours and $\alpha_k$ is the hyperparameter for the k-th hop. The whole loss function can be expressed as:
\begin{equation}
L_{NodeGAE} =  L_{LM} + L_{InfoNCE}.
\end{equation}
This approach allows the model to simultaneously learn the semantic information from the text data as well as the structural information in the graph. By optimizing both text reconstruction and neighborhood-based graph losses, the model can produce high-quality node embeddings that encode both textual and structural knowledge.
\\

\begin{table*}[t]
\centering
\large
\begin{tabular}{@{}llccccc@{}}
\toprule
\multicolumn{1}{l}{Dataset}  & Classifier & $h_{shallow}$ & $h_{sent-emb}$ & $h_{lm-finetune}$ & $h_{giant}$ & $h_{NodeGAE}$ (Ours)  \\ \midrule
\multirow{4}{*}{ogbn-arxiv}  & MLP              & 54.21 $\pm$ 0.23 & 69.60 $\pm$ 0.25 & 72.28 $\pm$ 0.14 & 73.08 $\pm$ 0.06 & 73.71 $\pm$ 0.10 \\                & GCN & 71.82 $\pm$ 0.21 & 73.21 $\pm$ 0.07 & 74.68 $\pm$ 0.18 & 73.29 $\pm$ 0.10 & 73.76 $\pm$ 0.08 \\
                             & GraphSAGE        & 71.13 $\pm$ 0.26 & 73.61 $\pm$ 0.13 & 74.43 $\pm$ 0.23 & 74.59 $\pm$ 0.28 & 75.38 $\pm$ 0.11 \\
                             & RevGAT  & 73.16 $\pm$ 0.09 & 75.20 $\pm$ 0.11 & 75.10 $\pm$ 0.09 & 76.12 $\pm$ 0.16 & \textbf{77.10 $\pm$ 0.08} \\ \midrule


\multirow{4}{*}{ogbn-products} & MLP              & 60.47 $\pm$ 0.45 & 77.22 $\pm$ 0.04 & 58.74 $\pm$ 0.24 & 77.58 $\pm$ 0.24 & 80.25 $\pm$ 0.23 \\
                                    & ClusterGCN        & 79.29 $\pm$ 0.14 & 83.44 $\pm$ 0.20 & 58.53 $\pm$ 0.45 & 82.84 $\pm$ 0.29 & 84.20  $\pm$ 0.22 \\ 
                                    & GAMLP              & 83.54 $\pm$ 0.09 & 84.59 $\pm$ 0.06 & 77.18 $\pm$ 0.78 & 83.16 $\pm$ 0.07 & 85.62 $\pm$ 0.12 \\
                                    & SAGN+SCR         & 77.50 $\pm$ 0.14 &  85.07 $\pm$ 0.30  & 76.78 $\pm$ 0.83 & 85.79 $\pm$ 0.14 
                                    & \textbf{86.32 $\pm$ 0.09} \\
                             


\bottomrule

\end{tabular}
\vspace{0.15cm}
\caption{Node classification performance on the ogbn-arxiv and ogbn-products. Results report the mean accuracy $\pm$ one standard deviation over 10 repeated runs. The best-performing methods are highlighted in bold.}
\label{table:node_cls}
\vspace{-0.3in}
\end{table*}

\begin{table}[h]
\begin{tabular}{lcc}
\toprule
\multicolumn{1}{l}{Method} & MLP & GraphSAGE \\ \hline
                   $h_{shallow}$ & 89.31 $\pm$ 0.06  & 96.85 $\pm$ 0.07 \\
                     $h_{sent-emb}$  & 96.58 $\pm$ 0.03  & 96.60 $\pm$ 0.11 \\
                     $h_{lm-finetune}$ & 97.30 $\pm$ 0.08 & 97.82 $\pm$ 0.10 \\
                   $h_{NodeGAE}$ (Ours)  & \textbf{99.39 $\pm$ 0.01} & 98.28 $\pm$ 0.06 \\ \bottomrule
\end{tabular}
\vspace{0.15cm}
\caption{The link prediction ROC-AUC results on the ogbn-arxiv dataset, which is created by us using random link sampling. The best result is highlighted in bold. }
\label{table:link_pred}
\vspace{-1cm}

\end{table}

\vspace{-0.2in}

\subsection{Variational Framework}
Our proposed approach, NodeGAE, can be formulated into a variational framework. Let $p_\theta(\mathcal{T}|\mathcal{V},\mathcal{E})$ represent the distribution over the text $\mathcal{T}$ of the corresponding node $\mathcal{V}$ with its edge $\mathcal{E}$ and $q_\phi(H|\mathcal{T},\mathcal{V},\mathcal{E})$ represent the estimated posterior distribution over the latent representation $H$ for data in the textual graph, where $\mathcal{H,T,V,E}$ are random variables, and $\theta, \phi$ represent the parameters for the encoder and the decoder. The goal is to maximize the distribution $p_\theta(\mathcal{T}|\mathcal{V},\mathcal{E})$. From the original Variational Autoencoder Encoder (VAE) framework~\cite{kingma2022autoencoder_cv}, we know that the distribution is intractable, yet it has an Evidence Lower Bound (ELBO) that can be used for optimization. We have also derived the ELBO for NodeGAE:
\begin{equation} \label{eq1}
\begin{split}
\log p_\theta(\mathcal{T|V,E}) \geq  &  \ \mathbb{E}_{h\sim q_\phi(H|\mathcal{T,V,E})}\ [\log p_\theta(\mathcal{T}|H,\mathcal{V,E})] \\
 & -D_{KL}(q_\phi(H|\mathcal{T,V,E})\ \| \ p_\theta(H|\mathcal{V,E})),
\end{split}
\end{equation}
where $D_{KL}(q\|p)$ represent the KL-divergence between distribution $q$ and $p$. For the first term on the right-hand side of the inequality, it represents the text reconstruction loss. For the second term, it means that the estimated posterior distribution $q_\phi(H|\mathcal{T,V,E})$ should be close to the prior distribution $p_\theta(H|\mathcal{V,E})$. The InfoNCE loss~\cite{oord2019infonce} used in NodeGAE encourages the encoder $\phi$ to learn the correlations among neighboring nodes, which implicitly forces the posterior and the prior distributions to be close. The ELBO optimization theoretically guarantees that NodeGAE can effectively learn a representation of the textual graph.

\vspace{-0.05in}
\subsection{Parameterization} The encoder and decoder in our autoencoder architecture are parameterized as a sentence embedding model~\cite{ni2021sentencet5, reimers2019sentencebert} and a T5-like~\cite{raffel2023t5} language model, \textit{i.e.} an encoder transformer model and an encoder-decoder transformer model, respectively. The feature embedding takes the average of the encoder outputs across all input tokens. To enhance the reconstruction performance, the embedding from the encoder is projected to a larger size embedding: $W_2\sigma(W_1h)$, where $W_1\in\mathbb R ^ {d_{enc}\times d_{enc}}, W_2\in\mathbb{R}^{sd_{dec}\times d_{enc}}$, $s$ is the sequence length, and $d_{enc}, d_{dec}$ are the embedding dimension of the encoder and decoder. The projected encoder embedding is then reshaped into a sequence of input-sized embeddings and fed as the input to the decoder: $Decoder(W_2\sigma(W_1h)).$ This reshaping step allows the decoder to process the encoded text representation in a sequence-to-sequence manner, generating the reconstructed text output. The decoder can more effectively learn the semantic information from an input sequence than from just an input embedding alone.


The parameterization design for NodeGAE allows the autoencoder to leverage the powerful text comprehension capabilities of language models for the text reconstruction task. By projecting the encoder embedding to a larger size, the model can better capture the nuanced semantic information in the input text, leading to improved reconstruction quality.
\\

\vspace{-0.15in}
\section{Experiments}
We have performed extensive experiments to evaluate the effectiveness of our proposed method NodeGAE, by showing the performance on downstream tasks: node classification and link prediction. In Section \ref{exp:setup}, we describe the experimental setup, including the datasets, models, and evaluation metrics used. In Section \ref{exp:result}, we present the main results of NodeGAE accross diverse models and textual graph datasets. In Section \ref{exp:ablation}, we conduct ablation studies to check the effectiveness of InfoNCE loss. Besides, in Section \ref{exp:convergence} we further investigate NodeGAE, demonstrating a faster convergence rate. Finally, in Section \ref{exp:reconstruction}, we demonstrate the text reconstruction process during the pretraining phrase.

\subsection{Experimental Setup}
\label{exp:setup}
\noindent \textbf{Datasets.} For node classification, we evaluate our method on two textual graph datasets: ogbn-arxiv and ogbn-products~\cite{hu2021opengraphbenchmarkdatasets}, where the details of the datasets are shown in Table \ref{table:data_statistics}. We keep the original split for the datasets, and the raw text data is provided by the officials.
For link prediction, we create a link prediction dataset based on ogbn-arxiv: we random sample the links in ogbn-arxiv for training, validating, and testing with a ratio of 7:2:1.\\

\vspace{-0.3cm}

\noindent \textbf{Autoencoder Models.} For the encoder, we use a sentence-T5-base model~\cite{ni2021sentencet5}, which is a T5-base encoder pretrained for text retrieval with 110M parameters. For the decoder, we use a T5-base model~\cite{raffel2023t5} with 223M parameters. The projection layer appended to the output layer of the encoder is a 2-layer MLP. The sequence length for the projection is $s=16$. We train the autoencoder using the Adam optimizer~\cite{kingma2017adam} with a learning rate of 1e-4 and 10,000 linear warm-up steps. The token sequence length is 256. For the InfoNCE loss, we sample the neighbors from 1-hop and 2-hop neighbors and set $\alpha_1=1, \alpha_2=0.1, \tau=0.5$ in Equation (\ref{eq:loss_neighbour}). \\

\begin{figure*}[!t]
    \centering
    \includegraphics[width=0.99\textwidth]{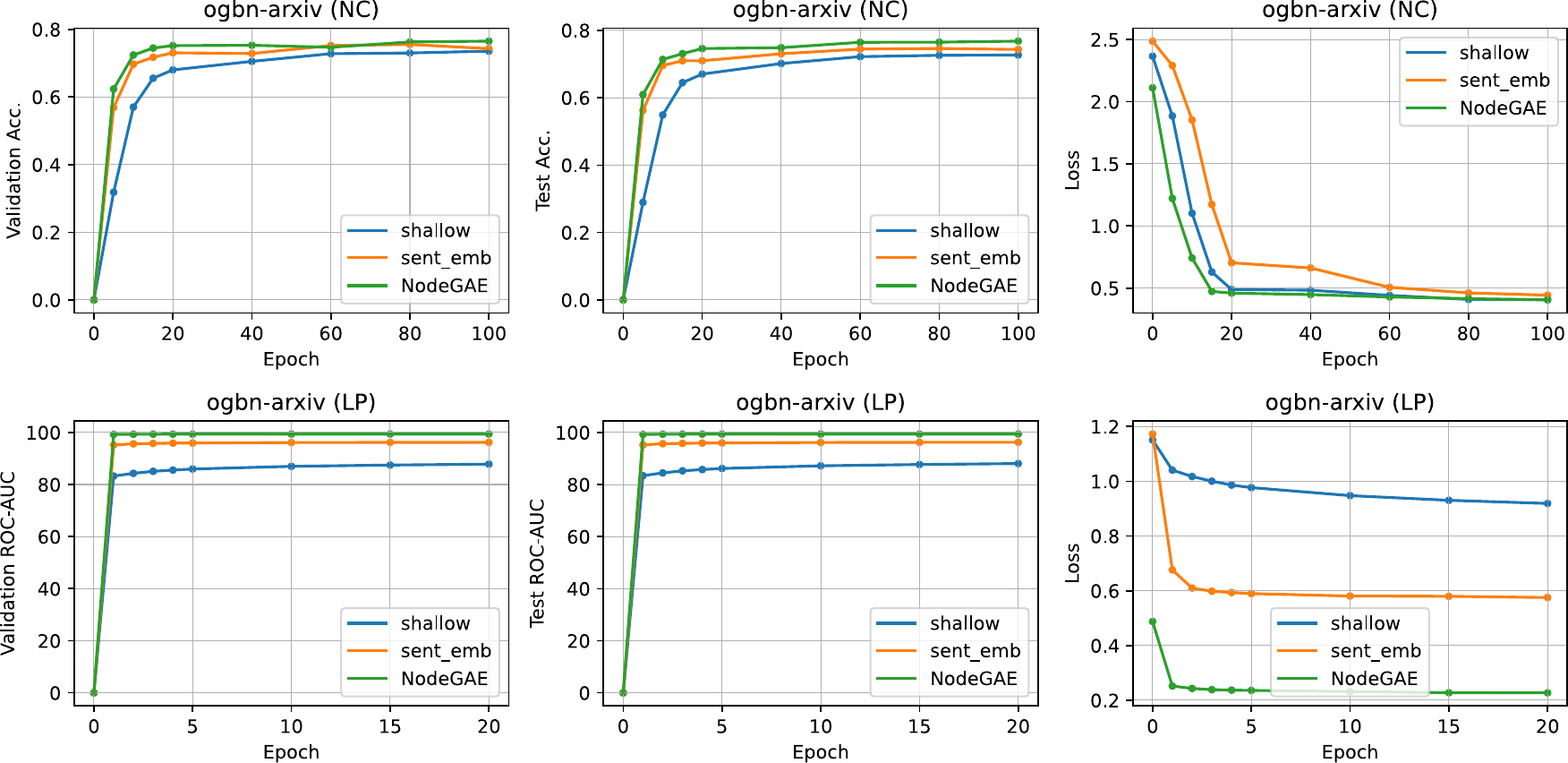}
    \vspace{-0.05in}
   \caption{To compare the convergence rates across different methods, we present the validation accuracy/ROC-AUC, test accuracy/ROC-AUC, and training loss curves for the training process on $h_{shallow}$, $h_{sent-emb}$, and $h_{NodeGAE}$ on the ogbn-arxiv dataset for node classification (NC) and link prediction (LP) tasks. The classifier for NC and LP are RevGAT and MLP respectively.}
    \label{fig:convergence}
    \vspace{-0.15in}
\end{figure*}

\vspace{-0.3cm}

\noindent \textbf{Classifiers.} We evaluate feature embeddings with commonly used baseline classifiers: MLP, GCN~\cite{Chiang2019clustergcn} and GraphSage~\cite{hamilton2018graphsage} to perform node classification and link prediction. Due to the vast scale of the ogbn-products dataset, training GCN and GraphSAGE becomes impractical. Therefore, we take ClusterGCN~\cite{Chiang2019clustergcn} as a substitution. Additionally, to achieve higher accuracy, we utilize SOTA GNN backbones: RevGAT~\cite{li2022revgat}, GAMLP~\cite{zhang2022gamlp} and SAGN+SCR~\cite{sun2021sagn, zhang2022scr}. We take the Adam optimizer~\cite{kingma2017adam} to train all the classifiers. For node classification, the learning rate is set to 1e-2 for MLP and GraphSAGE, 5e-3 for GAMLP and SAGN+SCR, 2e-3 for RevGAT; while for link prediction, the learning rate is set to 1e-4 for both MLP and GraphSAGE. 
\\


\vspace{-0.3cm}

\noindent \textbf{Evaluation Metrics.} For evaluation metrics, we use accuracy and the Area Under the ROC Curve (ROC-AUC)~\cite{bradley1997auc} for node classification and link prediction respectively. Accuracy measures the fraction of nodes that are correctly classified; ROC-AUC captures the model's ability to rank true links higher than false links. Both accuracy and ROC-AUC range from 0 to 1. We report the mean result $\pm$ one standard deviation over 10 repeated runs. The best-performing methods are highlighted in bold. \\

\vspace{-0.22in}

\subsection{Main Result}
\label{exp:result}
Following the settings described in \ref{exp:setup}, we evaluate NodeGAE by demonstrating the quality of the generated feature embeddings. The results are shown in Table \ref{table:node_cls} for node classification and in Table \ref{table:link_pred} for link prediction. To assess the quality, we evaluate the downstream performance on the feature embeddings, i.e., accuracy for node classification and ROC-AUC for link prediction. We compare the quality of our feature embeddings $h_{NodeGAE}$ with those generated by commonly used methods: $h_{shallow}$, shallow feature embeddings created by skip-gram~\cite{mikolov2013word2vec}; $h_{sent-emb}$, feature embeddings from a sentence-T5-base model~\cite{ni2021sentencet5}, a pretrained information retrieval model; $h_{lm-finetune}$, feature embeddings from a T5-base model finetuned on the text of labeled nodes; and $h_{giant}$, feature embeddings from a self-supervised learning method utilizing graph structure~\cite{chien2022giant}. The results in the tables show that our approach consistently outperforms the other methods across all datasets and models, demonstrating its effectiveness in improving textual graph learning. 


Furthermore, on ogbn-arxiv dataset for node classification, our best model (RevGAT) achieve an accuracy of 77.10\%, which is comparable to the performance of existing SOTA methods: SimTeG~\cite{duan2023simteg}, TAPE~\cite{he2024tape}, and GLEM~\cite{zhao2023glem}, which achieve an accuracy of 77.01\%, 77.50\%, and 76.94\%; for link prediction, our best model (MLP) reach a ROC-AUC score of 99.39\%, resulting in an improvement of 10.08\% over the performance of the shallow embedding $h_{shallow}$. Notably, on the ogbn-product dataset, which has a small scale of labeled training data (shown in Table \ref{table:data_statistics}), our best model (SAGN) achieved an accuracy of 86.32\%. This outperforms the supervised methods SimTeG and TAPE, which achieved accuracy of 85.40\% and 82.34\%; and the unsupervised method GIANT~\cite{chien2022giant} with an accuracy of 85.79\%. With the novel self-supervised learning autoencoder framework, NodeGAE can achieve superior performance when facing limited labeled training data. \\


\begin{table}[]
\begin{tabular}{@{}lccc@{}}
\toprule
\multicolumn{1}{l}{Datasets} & \#Nodes & \#Edges & Train/Val/Test \\ \midrule
                ogbn-arxiv     & 169,343 &  1,166,243 & 54/18/28 \\
                ogbn-products  & 2,449,029  & 61,859,140 & 8/2/90   \\
                 \bottomrule
\end{tabular}
\vspace{0.05in}
\caption{Statistics of the datasets.}
\label{table:data_statistics}
\vspace{-0.4in}
\end{table}

\begin{table}[h]
\begin{tabular}{@{}clc@{}}
\toprule
\multicolumn{1}{c}{} Rank & Method &  Test Acc. \\ \midrule
                    1 & NodeGAE (Ours) + SimTeG + TAPE  & 78.34 $\pm$ 0.06\\
                    2 & SimTeG + TAPE & 78.03 $\pm$ 0.07 \\
                    3 & NodeGAE (Ours) + TAPE  & 77.90 $\pm$ 0.10 \\
                    4 & TAPE~\cite{he2024tape}  & 77.50 $\pm$ 0.12 \\
                    5 & SimTeG~\cite{duan2023simteg} & 77.01 $\pm$ 0.13 \\
                    5 & GLEM~\cite{zhao2023glem}  & 76.94 $\pm$ 0.19 \\ \bottomrule
\end{tabular}
\vspace{0.05in}
\caption{We compare NodeGAE against existing SOTA methods on the ogbn-arxiv node classification task. We select the top-3 methods from the ogbn-arxiv leaderboard (accessed on 2024-08-04). The GNN baseline for all the results is RevGAT. }
\label{table:sota}
\vspace{-0.15in}
\end{table}

\vspace{-0.3cm}
\noindent \textbf{Compared with SOTAs.} We also compare our method with some existing SOTA approaches: SimTeG~\cite{duan2023simteg}, TAPE~\cite{he2024tape} and GLEM~\cite{zhao2023glem}, as shown in Table~\ref{table:sota}. On the ogbn-arxiv dataset, we achieve a new SOTA performance by ensembling 3 GNN models trained on feature embeddings from NodeGAE, SimTeG, and TAPE. Specifically, we use the embeddings officially provided by SimTeG for training a GNN. TAPE uses GPT-3.5 Turbo~\cite{ouyang2022traininglanguagemodelsfollow} to predict 5 possible classes for each node, and we take the 5-dimensional embeddings as feature embeddings to train another GNN. The ensembled GNNs reach a new SOTA accuracy of 78.34\% on the ogbn-arxiv dataset, surpassing the previous top result on the ogbn-arxiv leaderboard, which is achieved by the SimTeG+TAPE model with an accuracy of 78.03\%.

\vspace{-0.2cm}
\subsection{Ablation Study}
\label{exp:ablation}
We conduct ablation studies to show the extent to which the quality of the feature embeddings can be improved by using the InfoNCE loss~\cite{oord2019infonce}. The results are shown in Table \ref{table:ablation}. The experiment is performed on the node classification and link prediction using ogbn-arxiv dataset. $\Delta$ represents the margin of improved performance when using the InfoNCE loss. As demonstrated in the table, InfoNCE can generally enhance the performance of NodaGAE. Particularly for MLP, there is a 4.34\% improvement in accuracy for node classification and an 8.74\% enhancement in ROC-AUC for link prediction.

\begin{table}[h]
\small
\vspace{-0.3cm}
\begin{tabular}{@{}lclcccc@{}}
\toprule
\multicolumn{1}{c}{} &Metric &Method & w.o InfoNCE & w. InfoNCE & $\Delta$ \\ \midrule
\multirow{4}{*}{NC}    & \multirow{4}{*}{Acc.} & MLP & 69.37 $\pm$ 0.16 & 73.71 $\pm$ 0.10 & +4.34  \\
                     &   &  GCN & 73.38 $\pm$ 0.09 & 73.76 $\pm$ 0.08 & +0.38 \\
                     &  &  GraphSAGE & 73.49 $\pm$ 0.14  & 75.38 $\pm$ 0.11 & +1.89 \\
                     &  & RevGAT & 75.40 $\pm$ 0.27 & 77.10 $\pm$ 0.08 & +1.70  \\ \midrule
\multirow{2}{*}{LP}    & \multirow{2}{*}{ROC-AUC} &  MLP & 90.65 $\pm$ 0.10 & 99.39 $\pm$ 0.01 & +8.74   \\
                     &  &  GraphSAGE & 97.45 $\pm$ 0.04 & 99.28 $\pm$ 0.06 & +1.83 \\ \bottomrule
\end{tabular}
\vspace{0.05in}
\caption{Ablation studies to examine the impact of the InfoNCE loss on performance. The 'w.o' and 'w.' prefixes denote 'without' and 'with'; NC and LP represent node classification and link prediction. The symbol $\Delta$ represents the performance improvement.}
\label{table:ablation}
\vspace{-0.2in}
\end{table}

\begin{figure}[!h]
    \vspace{-0.5cm}
    \centering
    \includegraphics[width=0.44\textwidth]{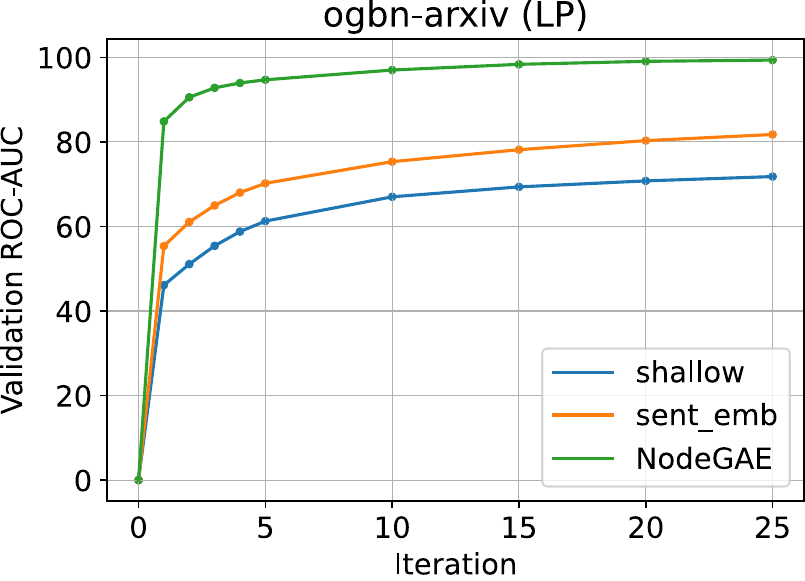}
    \vspace{-0.3cm}
    
   \caption{The validation ROC-AUC curve for link prediction during the first 25 iterations at the very beginning of the first epoch. We take MLP as the classifier.}
    \label{fig:convergence_2}
    \vspace{-0.15in}
\end{figure} 

\subsection{Convergence Analysis}
\label{exp:convergence}
We compare the convergence speed of the GNN trained on the feature embeddings of $h_{NodeGAE}$, $h_{shallow}$, and $h_{sent-emb}$, which is presented in Figure \ref{fig:convergence}. We illustrate the evolution of validation performance, test performance, and training loss over the course of increasing the number of training epochs for node classification and link prediction on the ogbn-arxiv dataset. As shown in the figure, the convergence speed of $h_{NodeGAE}$ is generally faster than that of $h_{shallow}$ and $h_{sent-emb}$, and it finally achieves the highest performance and lowest training loss. Notably, for link prediction, NodeGAE converges within 25 iterations at the very beginning of the first epoch, which is demonstrated in Figure \ref{fig:convergence_2}. Specifically, at the 25-th iteration, NodeGAE converges to a ROC-AUC score of 99.20\%; while the classifiers trained on the shallow features $h_{shallow}$ and sentence embedding features $h_{sent-emb}$ have margins of 18.54\% and 16.28\% respectively to reach the converged performance. \\

\subsection{Text Reconstruction}
\label{exp:reconstruction}
\begin{figure}[!h]
    \centering
    \includegraphics[width=0.4\textwidth]{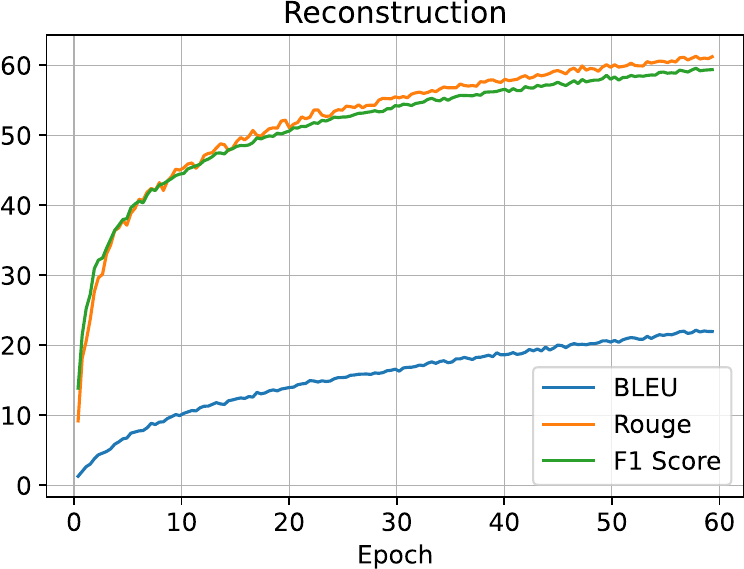}
    \vspace{-0.1in}
   \caption{NodeGAE text reconstruction process during pretraining phrase on ogbn-arxiv. }
    \label{fig:reconstruction}
    
\end{figure}

To test whether the autoencoder can successfully reconstruct the text, we show the BLEU~\cite{papineni2002bleu}, ROUGE~\cite{lin2004rouge}, and F1 scores of the generated text produced by the autoencoder during the pretraining stage, as illustrated in Figure \ref{fig:reconstruction}. The curve in the figure demonstrates that the autoencoder can reconstruct text with high quality. Specifically, the model can achieve BLEU, ROUGE, and F1 scores of 21.98\%, 61.20\%, and 59.36\%, respectively. These results indicate that the feature embeddings generated by the model contain rich textual information.

\section{Conclusion}
In this work, we have proposed NodeGAE, a novel node-level graph autoencoder framework for textual graph representation learning. Our simple and general approach leverages unsupervised learning through text reconstruction, which allows the encoder to effectively capture the valuable textual information from the graph nodes. To further enhance the embeddings, we incorporate InfoNCE loss to capture the graph structure. By taking advantage of unsupervised learning and leveraging the synergy between the text and the graph structure, our model is able to generate high-quality node embeddings that lead to superior performance on downstream tasks. Our comprehensive experimental evaluation demonstrates that NodeGAE achieves promising performance across diverse datasets and downstream tasks. We believe that the insights gained from this work will inspire further research in this important and growing area.

\balance
\bibliographystyle{acm}
\bibliography{sample-base}

\begin{thebibliography}{10}

\bibitem{bradley1997auc}
{\sc Bradley, A.~P.}
\newblock The use of the area under the roc curve in the evaluation of machine learning algorithms.
\newblock {\em Pattern Recognition 30}, 7 (1997), 1145--1159.

\bibitem{cai2017social_2}
{\sc Cai, C., He, R., and McAuley, J.}
\newblock Spmc: Socially-aware personalized markov chains for sparse sequential recommendation, 2017.

\bibitem{Chiang2019clustergcn}
{\sc Chiang, W.-L., Liu, X., Si, S., Li, Y., Bengio, S., and Hsieh, C.-J.}
\newblock Cluster-gcn: An efficient algorithm for training deep and large graph convolutional networks.
\newblock In {\em Proceedings of the 25th ACM SIGKDD International Conference on Knowledge Discovery \& Data Mining\/} (July 2019), KDD ’19, ACM.

\bibitem{chien2022giant}
{\sc Chien, E., Chang, W.-C., Hsieh, C.-J., Yu, H.-F., Zhang, J., Milenkovic, O., and Dhillon, I.~S.}
\newblock Node feature extraction by self-supervised multi-scale neighborhood prediction, 2022.

\bibitem{duan2023simteg}
{\sc Duan, K., Liu, Q., Chua, T.-S., Yan, S., Ooi, W.~T., Xie, Q., and He, J.}
\newblock Simteg: A frustratingly simple approach improves textual graph learning, 2023.

\bibitem{hamilton2018graphsage}
{\sc Hamilton, W.~L., Ying, R., and Leskovec, J.}
\newblock Inductive representation learning on large graphs, 2018.

\bibitem{hasanzadeh2020siggae}
{\sc Hasanzadeh, A., Hajiramezanali, E., Duffield, N., Narayanan, K.~R., Zhou, M., and Qian, X.}
\newblock Semi-implicit graph variational auto-encoders, 2020.

\bibitem{hassani2020mvgrl}
{\sc Hassani, K., and Khasahmadi, A.~H.}
\newblock Contrastive multi-view representation learning on graphs.
\newblock In {\em Proceedings of the 37th International Conference on Machine Learning\/} (13--18 Jul 2020), H.~D. III and A.~Singh, Eds., vol.~119 of {\em Proceedings of Machine Learning Research}, PMLR, pp.~4116--4126.

\bibitem{He_2016recsys_2}
{\sc He, R., and McAuley, J.}
\newblock Ups and downs: Modeling the visual evolution of fashion trends with one-class collaborative filtering.
\newblock In {\em Proceedings of the 25th International Conference on World Wide Web\/} (Apr. 2016), WWW ’16, International World Wide Web Conferences Steering Committee.

\bibitem{he2024tape}
{\sc He, X., Bresson, X., Laurent, T., Perold, A., LeCun, Y., and Hooi, B.}
\newblock Harnessing explanations: Llm-to-lm interpreter for enhanced text-attributed graph representation learning, 2024.

\bibitem{hu2021lora}
{\sc Hu, E.~J., Shen, Y., Wallis, P., Allen-Zhu, Z., Li, Y., Wang, S., Wang, L., and Chen, W.}
\newblock Lora: Low-rank adaptation of large language models, 2021.

\bibitem{hu2021opengraphbenchmarkdatasets}
{\sc Hu, W., Fey, M., Zitnik, M., Dong, Y., Ren, H., Liu, B., Catasta, M., and Leskovec, J.}
\newblock Open graph benchmark: Datasets for machine learning on graphs, 2021.

\bibitem{hu2020strategiespretraininggraphneural}
{\sc Hu, W., Liu, B., Gomes, J., Zitnik, M., Liang, P., Pande, V., and Leskovec, J.}
\newblock Strategies for pre-training graph neural networks, 2020.

\bibitem{hu2020gptgnn}
{\sc Hu, Z., Dong, Y., Wang, K., Chang, K.-W., and Sun, Y.}
\newblock Gpt-gnn: Generative pre-training of graph neural networks, 2020.

\bibitem{kingma2017adam}
{\sc Kingma, D.~P., and Ba, J.}
\newblock Adam: A method for stochastic optimization, 2017.

\bibitem{kingma2022autoencoder_cv}
{\sc Kingma, D.~P., and Welling, M.}
\newblock Auto-encoding variational bayes, 2022.

\bibitem{kipf2016gae}
{\sc Kipf, T.~N., and Welling, M.}
\newblock Variational graph auto-encoders, 2016.

\bibitem{kipf2017gcn}
{\sc Kipf, T.~N., and Welling, M.}
\newblock Semi-supervised classification with graph convolutional networks, 2017.

\bibitem{li2021adsgnn}
{\sc Li, C., Pang, B., Liu, Y., Sun, H., Liu, Z., Xie, X., Yang, T., Cui, Y., Zhang, L., and Zhang, Q.}
\newblock Adsgnn: Behavior-graph augmented relevance modeling in sponsored search, 2021.

\bibitem{li2022revgat}
{\sc Li, G., Müller, M., Ghanem, B., and Koltun, V.}
\newblock Training graph neural networks with 1000 layers, 2022.

\bibitem{lin2004rouge}
{\sc Lin, C.-Y.}
\newblock {ROUGE}: A package for automatic evaluation of summaries.
\newblock In {\em Text Summarization Branches Out\/} (Barcelona, Spain, July 2004), Association for Computational Linguistics, pp.~74--81.

\bibitem{mcauley2013social}
{\sc McAuley, J., and Leskovec, J.}
\newblock Discovering social circles in ego networks, 2013.

\bibitem{mcauley2015recsys}
{\sc McAuley, J., Targett, C., Shi, Q., and van~den Hengel, A.}
\newblock Image-based recommendations on styles and substitutes, 2015.

\bibitem{mikolov2013word2vec}
{\sc Mikolov, T., Sutskever, I., Chen, K., Corrado, G., and Dean, J.}
\newblock Distributed representations of words and phrases and their compositionality, 2013.

\bibitem{ni2021sentencet5}
{\sc Ni, J., Ábrego, G.~H., Constant, N., Ma, J., Hall, K.~B., Cer, D., and Yang, Y.}
\newblock Sentence-t5: Scalable sentence encoders from pre-trained text-to-text models, 2021.

\bibitem{ouyang2022traininglanguagemodelsfollow}
{\sc Ouyang, L., Wu, J., Jiang, X., Almeida, D., Wainwright, C.~L., Mishkin, P., Zhang, C., Agarwal, S., Slama, K., Ray, A., Schulman, J., Hilton, J., Kelton, F., Miller, L., Simens, M., Askell, A., Welinder, P., Christiano, P., Leike, J., and Lowe, R.}
\newblock Training language models to follow instructions with human feedback, 2022.

\bibitem{papineni2002bleu}
{\sc Papineni, K., Roukos, S., Ward, T., and Zhu, W.-J.}
\newblock {B}leu: a method for automatic evaluation of machine translation.
\newblock In {\em Proceedings of the 40th Annual Meeting of the Association for Computational Linguistics\/} (Philadelphia, Pennsylvania, USA, July 2002), P.~Isabelle, E.~Charniak, and D.~Lin, Eds., Association for Computational Linguistics, pp.~311--318.

\bibitem{paszke2019pytorch}
{\sc Paszke, A., Gross, S., Massa, F., Lerer, A., Bradbury, J., Chanan, G., Killeen, T., Lin, Z., Gimelshein, N., Antiga, L., Desmaison, A., Köpf, A., Yang, E., DeVito, Z., Raison, M., Tejani, A., Chilamkurthy, S., Steiner, B., Fang, L., Bai, J., and Chintala, S.}
\newblock Pytorch: An imperative style, high-performance deep learning library, 2019.

\bibitem{raffel2023t5}
{\sc Raffel, C., Shazeer, N., Roberts, A., Lee, K., Narang, S., Matena, M., Zhou, Y., Li, W., and Liu, P.~J.}
\newblock Exploring the limits of transfer learning with a unified text-to-text transformer, 2023.

\bibitem{reimers2019sentencebert}
{\sc Reimers, N., and Gurevych, I.}
\newblock Sentence-bert: Sentence embeddings using siamese bert-networks, 2019.

\bibitem{rong2020grover}
{\sc Rong, Y., Bian, Y., Xu, T., Xie, W., Wei, Y., Huang, W., and Huang, J.}
\newblock Self-supervised graph transformer on large-scale molecular data, 2020.

\bibitem{sap2019KG_2}
{\sc Sap, M., LeBras, R., Allaway, E., Bhagavatula, C., Lourie, N., Rashkin, H., Roof, B., Smith, N.~A., and Choi, Y.}
\newblock Atomic: An atlas of machine commonsense for if-then reasoning, 2019.

\bibitem{shen2020autoencoder_nlp}
{\sc Shen, T., Mueller, J., Barzilay, R., and Jaakkola, T.}
\newblock Educating text autoencoders: Latent representation guidance via denoising, 2020.

\bibitem{speer2018KG}
{\sc Speer, R., Chin, J., and Havasi, C.}
\newblock Conceptnet 5.5: An open multilingual graph of general knowledge, 2018.

\bibitem{sun2021sagn}
{\sc Sun, C., Gu, H., and Hu, J.}
\newblock Scalable and adaptive graph neural networks with self-label-enhanced training, 2021.

\bibitem{sun2020infograph}
{\sc Sun, F.-Y., Hoffmann, J., Verma, V., and Tang, J.}
\newblock Infograph: Unsupervised and semi-supervised graph-level representation learning via mutual information maximization, 2020.

\bibitem{susheel2021advance}
{\sc Suresh, S., Li, P., Hao, C., and Neville, J.}
\newblock Adversarial graph augmentation to improve graph contrastive learning.
\newblock In {\em Advances in Neural Information Processing Systems\/} (2021), M.~Ranzato, A.~Beygelzimer, Y.~Dauphin, P.~Liang, and J.~W. Vaughan, Eds., vol.~34, Curran Associates, Inc., pp.~15920--15933.

\bibitem{tan2022mgae}
{\sc Tan, Q., Liu, N., Huang, X., Chen, R., Choi, S.-H., and Hu, X.}
\newblock Mgae: Masked autoencoders for self-supervised learning on graphs, 2022.

\bibitem{oord2019infonce}
{\sc van~den Oord, A., Li, Y., and Vinyals, O.}
\newblock Representation learning with contrastive predictive coding, 2019.

\bibitem{velickovic2018DGI}
{\sc Veličković, P., Fedus, W., Hamilton, W.~L., Liò, P., Bengio, Y., and Hjelm, R.~D.}
\newblock Deep graph infomax, 2018.

\bibitem{wang2020MAG}
{\sc Wang, K., Shen, Z., Huang, C., Wu, C.-H., Dong, Y., and Kanakia, A.}
\newblock {Microsoft Academic Graph: When experts are not enough}.
\newblock {\em Quantitative Science Studies 1}, 1 (02 2020), 396--413.

\bibitem{xia2022progclrethinkinghardnegative}
{\sc Xia, J., Wu, L., Wang, G., Chen, J., and Li, S.~Z.}
\newblock Progcl: Rethinking hard negative mining in graph contrastive learning, 2022.

\bibitem{yang2016revisitingsemisupervisedlearninggraph}
{\sc Yang, Z., Cohen, W.~W., and Salakhutdinov, R.}
\newblock Revisiting semi-supervised learning with graph embeddings, 2016.

\bibitem{you2021graphcontrastivelearningautomated}
{\sc You, Y., Chen, T., Shen, Y., and Wang, Z.}
\newblock Graph contrastive learning automated, 2021.

\bibitem{you2021graphcl}
{\sc You, Y., Chen, T., Sui, Y., Chen, T., Wang, Z., and Shen, Y.}
\newblock Graph contrastive learning with augmentations, 2021.

\bibitem{zhang2022scr}
{\sc Zhang, C., He, Y., Cen, Y., Hou, Z., Feng, W., Dong, Y., Cheng, X., Cai, H., He, F., and Tang, J.}
\newblock Scr: Training graph neural networks with consistency regularization, 2022.

\bibitem{zhang2022gamlp}
{\sc Zhang, W., Yin, Z., Sheng, Z., Li, Y., Ouyang, W., Li, X., Tao, Y., Yang, Z., and Cui, B.}
\newblock Graph attention multi-layer perceptron.
\newblock In {\em Proceedings of the 28th ACM SIGKDD Conference on Knowledge Discovery and Data Mining\/} (Aug. 2022), KDD ’22, ACM.

\bibitem{zhang2021MGSSL}
{\sc Zhang, Z., Liu, Q., Wang, H., Lu, C., and Lee, C.-K.}
\newblock Motif-based graph self-supervised learning for molecular property prediction, 2021.

\bibitem{zhao2023glem}
{\sc Zhao, J., Qu, M., Li, C., Yan, H., Liu, Q., Li, R., Xie, X., and Tang, J.}
\newblock Learning on large-scale text-attributed graphs via variational inference, 2023.

\bibitem{zhu2021textgnn}
{\sc Zhu, J., Cui, Y., Liu, Y., Sun, H., Li, X., Pelger, M., Yang, T., Zhang, L., Zhang, R., and Zhao, H.}
\newblock Textgnn: Improving text encoder via graph neural network in sponsored search.
\newblock In {\em Proceedings of the Web Conference 2021\/} (Apr. 2021), WWW ’21, ACM.

\bibitem{zhu2020grace}
{\sc Zhu, Y., Xu, Y., Yu, F., Liu, Q., Wu, S., and Wang, L.}
\newblock Deep graph contrastive representation learning, 2020.

\bibitem{Zhu_2021}
{\sc Zhu, Y., Xu, Y., Yu, F., Liu, Q., Wu, S., and Wang, L.}
\newblock Graph contrastive learning with adaptive augmentation.
\newblock In {\em Proceedings of the Web Conference 2021\/} (Apr. 2021), WWW ’21, ACM.

\end{thebibliography}

\end{document}